\documentclass[conference]{IEEEtran}
\IEEEoverridecommandlockouts
\usepackage{cite}
\usepackage{amsmath,amssymb,amsfonts}
\usepackage{algorithmic}
\usepackage{graphicx}
\usepackage{textcomp}
\usepackage{xcolor}
\usepackage[hang,flushmargin]{footmisc}
\usepackage{subfigure}
\usepackage{booktabs}

\def\BibTeX{{\rm B\kern-.05em{\sc i\kern-.025em b}\kern-.08em
    T\kern-.1667em\lower.7ex\hbox{E}\kern-.125emX}}
\begin{document}

\title{Integrated Drill Boom Hole-Seeking Control via Reinforcement Learning\\
}

\author{\IEEEauthorblockN{Haoqi Yan}
\IEEEauthorblockA{\textit{School of Mechanical Engineering} \\
\textit{University of Science and}\\
\textit{Technology Beijing}\\
Beijing, China \\
yhq0925@xs.ustb.edu.cn}
\and
\IEEEauthorblockN{Haoyuan Xu}
\IEEEauthorblockA{\textit{School of Mechanical Engineering} \\
\textit{University of Science and}\\
\textit{Technology Beijing}\\
Beijing, China \\
xhy@xs.ustb.edu.cn}
\and
\IEEEauthorblockN{Hongbo Gao}
\IEEEauthorblockA{\textit{School  of Information Science and Technology} \\
\textit{University of Science and}\\
\textit{Technology of China}\\
Hefei, China \\
ghb48@ustc.edu.cn}
\and
\IEEEauthorblockN{Fei Ma}
\IEEEauthorblockA{\textit{School of Mechanical Engineering} \\
\textit{University of Science and}\\
\textit{Technology Beijing}\\
Beijing, China \\
yeke@ustb.edu.cn}
\and
\IEEEauthorblockN{Shengbo Eben Li}
\IEEEauthorblockA{\textit{School of Vehicle and Mobility} \\
\textit{Tsinghua University}\\
Beijing, China \\
lishbo@tsinghua.edu.cn}
\and
\IEEEauthorblockN{Jingliang Duan*}
\IEEEauthorblockA{\textit{School of Mechanical Engineering} \\
\textit{University of Science and}\\
\textit{Technology Beijing}\\
Beijing, China \\
duanjl@ustb.edu.cn \\
}}

\maketitle
\renewcommand{\footnoterule}{\vspace*{-3pt}\hrule width \columnwidth \vspace*{2pt}} 
\let\thefootnote\relax\footnotetext{\noindent*Corresponding Author\\The work was supported in part by the NSF China under Grant 52202487 and in part by the State Key Laboratory of Automotive Safety and Energy, China under Project KFY2212. }
\begin{abstract}
Intelligent drill boom hole-seeking is a promising technology for enhancing drilling efficiency, mitigating potential safety hazards, and relieving human operators. Most existing intelligent drill boom control methods rely on a hierarchical control framework based on inverse kinematics. However, these methods are generally time-consuming due to the computational complexity of inverse kinematics and the inefficiency of the sequential execution of multiple joints. To tackle these challenges, this study proposes an integrated drill boom control method based on Reinforcement Learning (RL). We develop an integrated drill boom control framework that utilizes a parameterized policy to directly generate control inputs for all joints at each time step, taking advantage of joint posture and target hole information. By formulating the hole-seeking task as a Markov decision process, contemporary mainstream RL algorithms can be directly employed to learn a hole-seeking policy, thus eliminating the need for inverse kinematics solutions and promoting cooperative multi-joint control. To enhance the drilling accuracy throughout the entire drilling process, we devise a state representation that combines Denavit-Hartenberg joint information and preview hole-seeking discrepancy data. Simulation results show that the proposed method significantly outperforms traditional methods in terms of hole-seeking accuracy and time efficiency.
\end{abstract}

\begin{IEEEkeywords}
reinforcement learning, integrated drill boom control, hole seeking, robotic arm
\end{IEEEkeywords}
\section{Introduction}

The rock drilling jumbo holds immense significance in underground drilling and tunneling operations due to its exceptional adaptability and cost advantages\cite{cui2020positioning,liu2016finite}. As the core component of the jumbo, the drill boom plays a crucial role in the hole-seeking operation\cite{li2018intelligent}, which serves as the fundamental basis of rock drilling. In practical applications, controlling the boom to achieve the desired posture manually is extremely challenging and time-consuming due to its high degree of freedom (DOF). Additionally, the harsh underground conditions pose significant risks of accidents for human operators\cite{onifade2023safer}. To enhance working efficiency and safety, there is an urgent need to advance research on intelligent hole-seeking control technology, with the aim of replacing labor-intensive manual production operations\cite{zhao2023deep}.

\begin{figure}[htbp] 
\centering 
\includegraphics[width=0.4\textwidth]{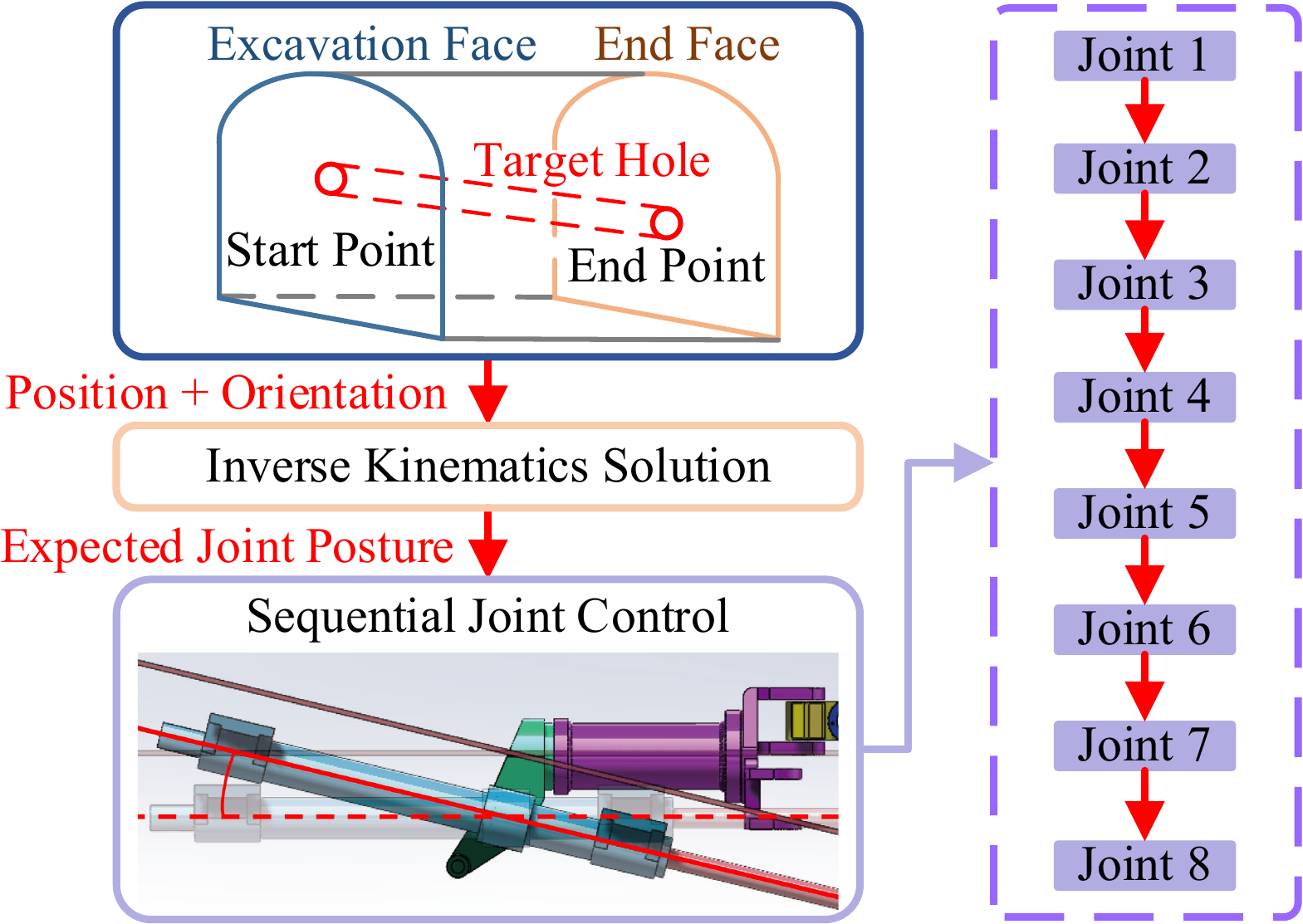}
\caption{Hierarchical framework of drill boom control.} 
\label{fig.hierarchical_framework} 
\end{figure}

Existing studies on intelligent hole-seeking control typically employ a hierarchical control framework, as depicted in Fig. \ref{fig.hierarchical_framework}\cite{navarro2018mutual,wang2015automatic}. This framework consists of two layers: the inverse-kinematics solution layer and the sequential joint control layer\cite{fodera2020factors}. The upper layer determines the expected posture for each joint by considering the position and orientation of the target hole, thereby ensuring the alignment of the drill boom end with the start point of the target hole and facilitating drilling along the expected direction. Subsequently, the lower layer controls each joint individually, guiding them to transition from their current states to the calculated expected postures following a predefined sequence\cite{cardu2016quantifying,cui2020positioning}.

In the upper layer, the current mainstream inverse-kinematics methods can be broadly classified into two categories: analytic methods and numerical methods. Analytic methods can provide accurate and efficient inverse-kinematics solutions, but they are generally not suitable for boom systems with high DOF  (DOF$>$6)\cite{ju2018inverse}.
Numerical methods, on the other hand, have the potential to handle high-DOF boom systems but often exhibit lower solution speeds and  accuracy\cite{wang2015inverse}. Moreover, since the drill boom is a redundant structure, meaning that multiple feasible solutions may exist for inverse kinematics \cite{alkayyali2019pso}, neither analytic nor numerical methods can guarantee the discovery of the optimal solution from the feasible solution set \cite{kucuk2014inverse}. In the lower layer, each joint is individually controlled to reach its target position in a sequential manner, due to the absence of a multi-joint coordination mechanism in the hierarchical framework. This tends to be less time-efficient compared to cooperative control across multiple joints\cite{hernandez2021inverse}.

In summary, the hierarchical drill boom control framework based on inverse kinematics typically suffers from low execution efficiency. Inspired by the considerable success of reinforcement learning (RL) in robot control\cite{duan2023relaxed,li2023reinforcement,gao2019research,wang2017parallel}, this study proposes an RL-based integrated control method for drill boom hole-seeking. The main contribution is summarized as follows:
\begin{enumerate}
    \item We establish an integrated drill boom control framework that utilizes a parameterized policy to directly generate control inputs for all joints at each time step, leveraging joint posture and target hole information. Building upon this integrated framework design, we formulate the hole-seeking task as a Markov Decision Process (MDP) and employ RL to learn a hole-seeking control policy. In comparison to the hierarchical drill boom control framework, our method eliminates the need for inverse kinematics solutions and enables cooperative multi-joint control, significantly reducing drilling time consumption.
    \item To further improve control performance, we devise a state representation that encompasses two types of information: (1) the 8-dimensional joint posture information in the Denavit-Hartenberg coordinate system, and (2) the current and preview hole-seeking errors between the drill end and the target hole. Compared to representations based on Cartesian coordinates and non-preview hole-seeking errors, our characterization exhibits notably superior hole-seeking accuracy.
\end{enumerate}

The remainder of this paper is organized as follows: 
Section \ref{sec.priliminary} provides an introduction to the prior knowledge utilized in the subsequent sections. Section \ref{sec.integrated framework} proposes the integrated framework for drill boom control. Section \ref{sec.MDP hole-seeking} focuses on formulating the essential elements for RL-based policy learning. Section \ref{sec.experiment} presents numerical results and Section \ref{sec.conclusion} concludes the paper.

\section{Preliminary}
\label{sec.priliminary}
\subsection{Reinforcement learning}
Reinforcement learning (RL) seeks to find the optimal policy of the Markov Decision Process (MDP), which is defined by a state space $\mathcal{S}$, an action space $\mathcal{A}$, transition dynamics $p(s_{t+1}|s_t, a_t)$, and a reward function $r(s_t,a_t): \mathcal{S} \times \mathcal{A} \rightarrow \mathbb{R}$.
At each time step $t$, the system observes a state $s_{t}$, takes an action $a_t$, receives a scalar reward $r$, and transitions to the next state $s_{t+1}$.
The action is selected according to the stochastic policy $\pi (a_t|s_t): \mathcal{S} \rightarrow \mathcal{P(A)}$, which maps a given state to  a probability distribution over $\mathcal{A}$.

The policy update goal of RL is to maximize the expected return for each state $s_t$. In particular, the return $R_t = \Sigma_{k=0}^\infty {\gamma^k} r_{t+k}$ represents the sum of the discounted future rewards from time step $t$, where $\gamma \in (0, 1]$ is a discount factor that balances the importance of immediate and future rewards. The expected return when following policy $\pi$ from state $s$ is often referred to as the value function, denoted as ${V_\pi}(s)$. 

\subsection{Hole-seeking task of drill boom}

\begin{figure}[!htbp] 
\centering 
\subfigure[Structure of 8-DOF drill boom.]{
\includegraphics[width=0.45\textwidth]{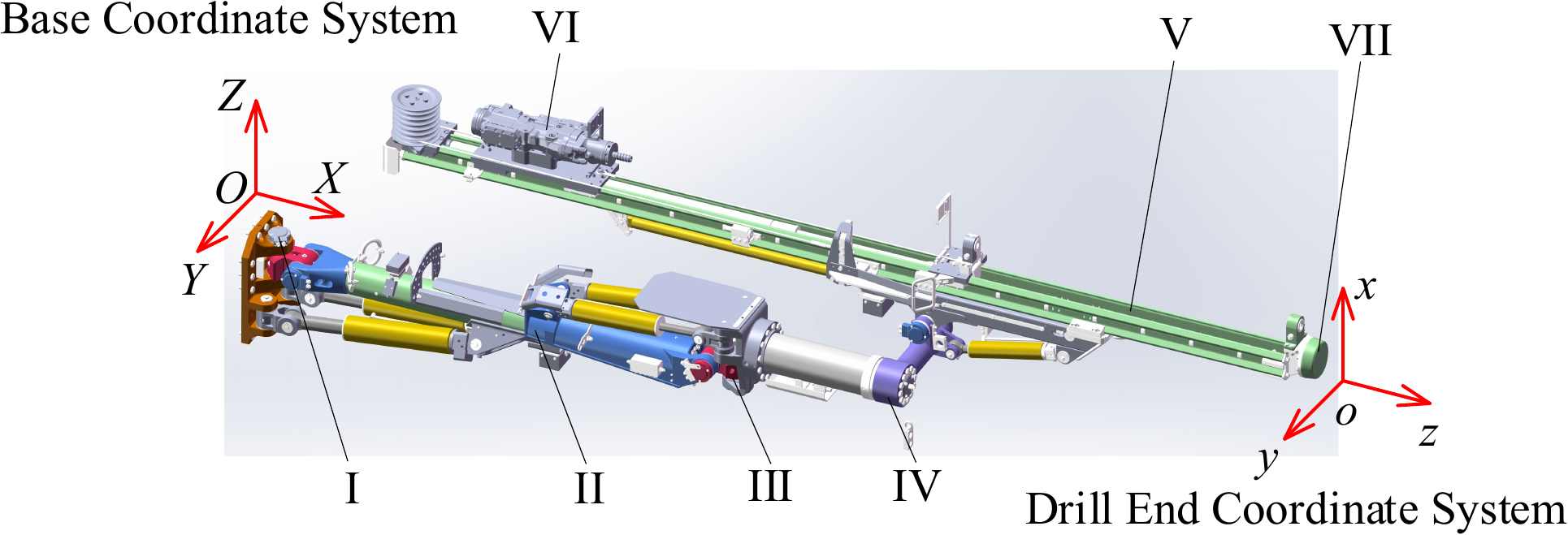}
\label{fig.structure}}\\
\subfigure[Simplified structure of 8-DOF drill boom.]{
\includegraphics[width=0.3\textwidth]{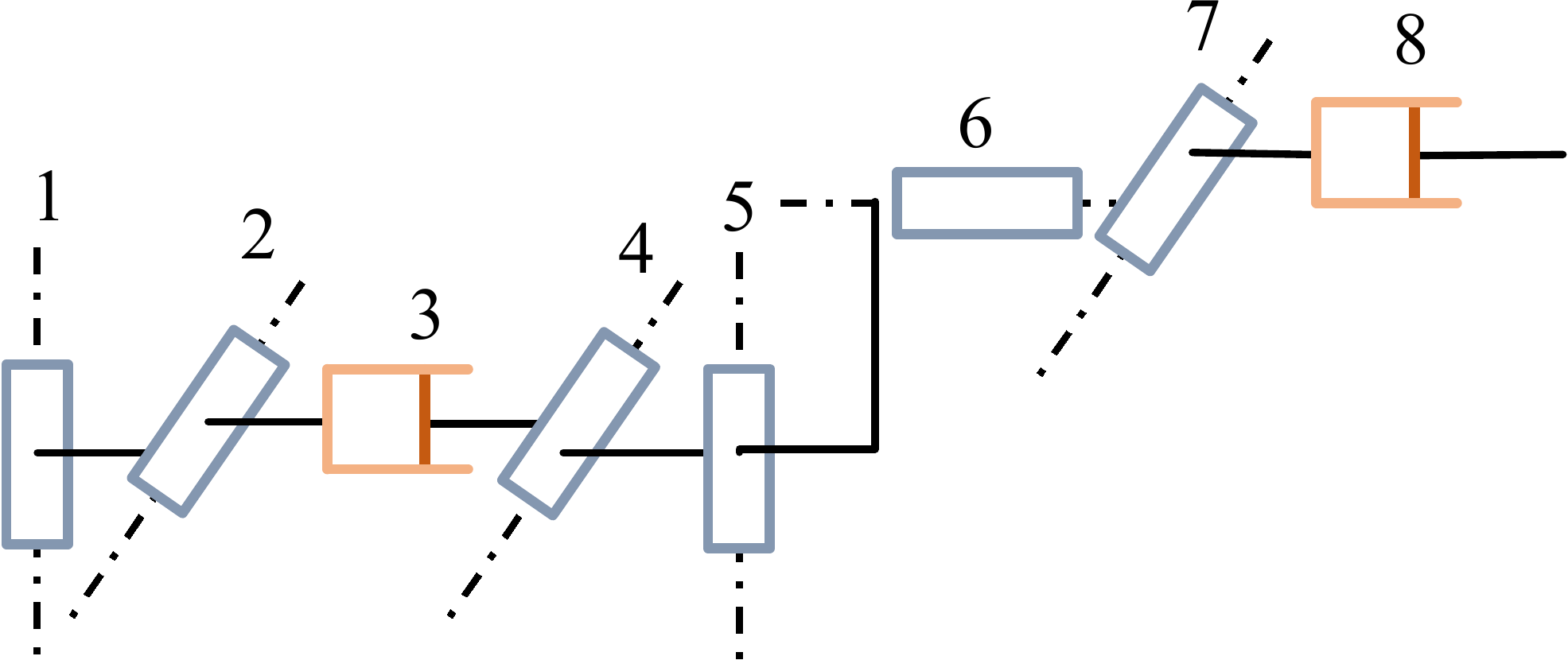}
\label{fig.diagram}}\\
\subfigure[Hole-seeking task description.]{
\includegraphics[width=0.45\textwidth]{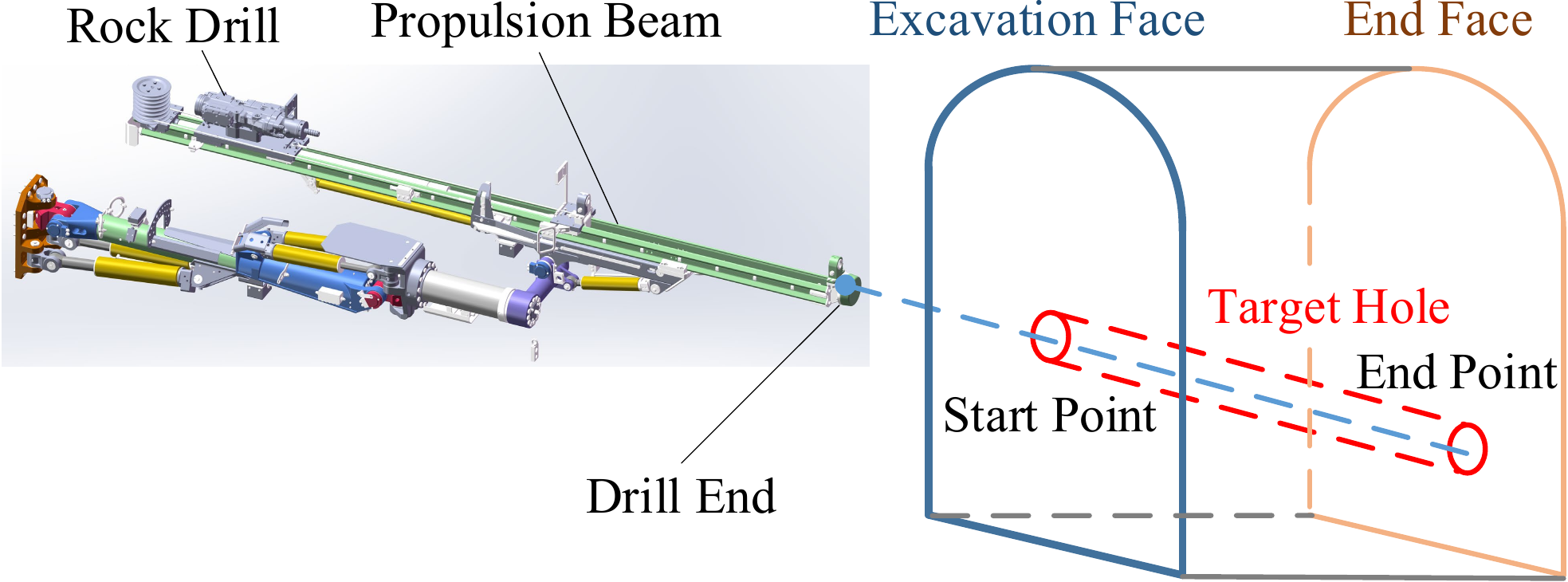}
\label{fig.task}}
\caption{Illustration of the 8-DOF drill boom and the hole-seeking task. I: installation seat; II: telescopic boom; III: rotary boom; IV: slewing boom; V: propulsion beam; VI: rock drill; VII: drill end.}
\label{fig.task_structure}
\end{figure}

This paper focuses on the 8-DOF drill boom depicted in Fig. \ref{fig.structure}. The drill boom can be simplified to the structure shown in Fig. \ref{fig.diagram}.
In this configuration, the 3rd and 8th joints are prismatic pairs, while the remaining joints are revolute pairs. Fig. \ref{fig.task} illustrates the objective of the hole-seeking task, which aims to find a controller that moves each joint to a desired posture, aligning the drill end with the start point of the target hole. Moreover, the orientation of the drill end should also match the direction of the target hole.

\subsection{Denavit-Hartenberg (DH) method}

In order to determine whether the drill end has reached the target point, it is necessary to calculate its position and orientation based on the current joint posture information. Therefore, the forward kinematics model of the drill boom is crucial. In this study, we utilize the Denavit-Hartenberg (DH) method to model the kinematics of the drill boom system\cite{denavit1955kinematic}.

In the DH method, each joint of the robot is associated with a coordinate frame. It employs a homogeneous transformation matrix to represent the translation and rotation between the coordinate frames of two adjacent joints. The transformation matrix incorporates four key parameters: joint angle $\theta$, link twist angle $\alpha$, link length $a$, and link offset $d$, which is expressed as
\begin{equation}
\label{eq.DH-T}
\setlength{\arraycolsep}{1.2pt}
T_{i-1}^{i} = \begin{bmatrix} \cos{\theta_i} & -\sin{\theta_i} \cos{\alpha_i} & \sin{\theta_i} \sin{\alpha_i} & a_{i-1} \cos{\theta_i} \\ \sin{\theta_i} & \cos{\theta_i} \cos{\alpha_i} & -\cos{\theta_i} \sin{\alpha_i} & a_{i-1} \sin{\theta_i} \\ 0 & \sin{\alpha_i} & \cos{\alpha_i} & d_i \\ 0 & 0 & 0 & 1 \end{bmatrix},
\end{equation}
where \(T_{i-1}^{i}\) represents the transformation matrix from frame $i$-$1$ to frame \(i\), \(\theta_i\) is the joint angle of the \(i\)th joint, \(\alpha_i\) is the link twist angle between the $i$-$1$th and \(i\)th frames, \(a_{i-1}\) is the link length of the $i$-$1$th frames, and \(d_i\) is the link offset between the $i$-$1$th and \(i\)th frames which is commonly employed to indicate the joint elongation of the prismatic pair.

By cascading the transformation matrices for all the joints, the forward kinematics of the drill boom system can be computed. This allows us to determine the position and orientation of the drill end based on the joint angles and elongations. Thus, from  \eqref{eq.DH-T}, the posture information of the drill end can be derived as
\begin{equation}
\nonumber
\begin{aligned}
T_{0}^{8} &= {T_{0}^{1}}{T_{1}^{2}}{T_{2}^{3}}{T_{3}^{4}}{T_{4}^{5}}{T_{5}^{6}}{T_{6}^{7}}{T_{7}^{8}}= \begin{bmatrix} R_{11} & R_{12} & R_{13} & x_{\rm drill} \\ 
R_{21} & R_{22} & R_{23} & y_{\rm drill} \\ R_{31} & R_{32} & R_{33} & z_{\rm drill} \\ 
0 & 0 & 0 & 1 \end{bmatrix}.
\end{aligned}
\end{equation}

The matrix $T_{0}^{8}$ can be split into two components\cite{paul1979kinematic}. The upper left \(3 \times 3\) elements reveal the rotational transformation relationship between the drill end and the base coordinate system. This information can be utilized to obtain the direction vector of the drill end in the base coordinate system. On the other hand, the coordinates $(x_{\rm drill},y_{\rm drill},z_{\rm drill})$ on the right-hand side directly reflect the position of the actual drill end in the base coordinate system.

\section{Integrated drill boom Control Framework}
\label{sec.integrated framework}

\begin{figure}[htbp] 
\centering 
\includegraphics[width=0.4\textwidth]{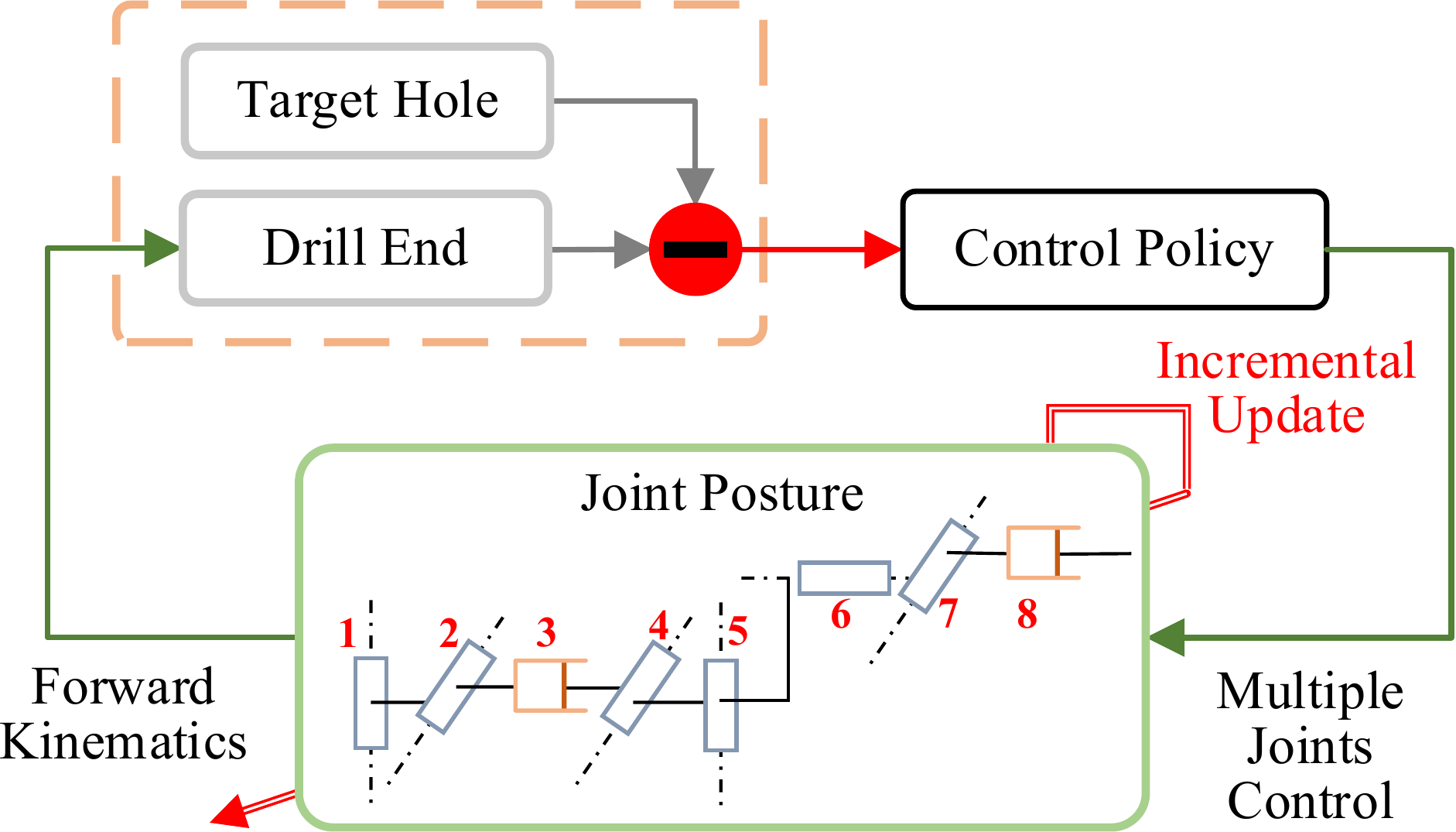}
\caption{Integrated framework of drill boom control.}  
\label{fig.framework}
\end{figure}

This paper proposes an integrated drill boom control framework. As shown in Fig.~\ref{fig.framework}, the joint posture is calculated by forward kinematics to obtain the posture information of the drill end. Leveraging the joint posture and the discrepancy between the drill end and the target hole, the control policy directly produces control signals for all joints at each time step and obtains the next-frame joint posture. This framework differs from the traditional hierarchical framework showcased in Fig.~\ref{fig.hierarchical_framework} in three significant ways: 
 \begin{enumerate}
     \item It eliminates the need for solving inverse kinematics, thus significantly enhancing the hole-seeking efficiency. 
    \item Instead of  controlling each joint sequentially, the integrated policy coordinates multiple joints simultaneously, further improving control efficiency.
    \item The policy can adaptively adjust the control inputs for each joint according to the hole-seeking error during the drilling process, which improves control robustness.
 \end{enumerate}

For the effective implementation of integrated decision control, the critical task lies in obtaining the control policy. However, due to the significant non-linearity of the forward kinematics, calculating the control inputs for multiple joints in real time with limited computational resources is impractical. As a result, this research adopts a two-step approach: using RL to solve a parameterized policy offline, and then applying it online during the hole-seeking process. The prerequisite for employing RL to solve the drill boom control policy is to formulate the hole-seeking task as an MDP, which will be detailed in the subsequent section.

\section{MDP formulation of drill boom hole-seeking}
\label{sec.MDP hole-seeking}
\begin{figure}[!htbp] 
\centering 
\includegraphics[width=0.4\textwidth]{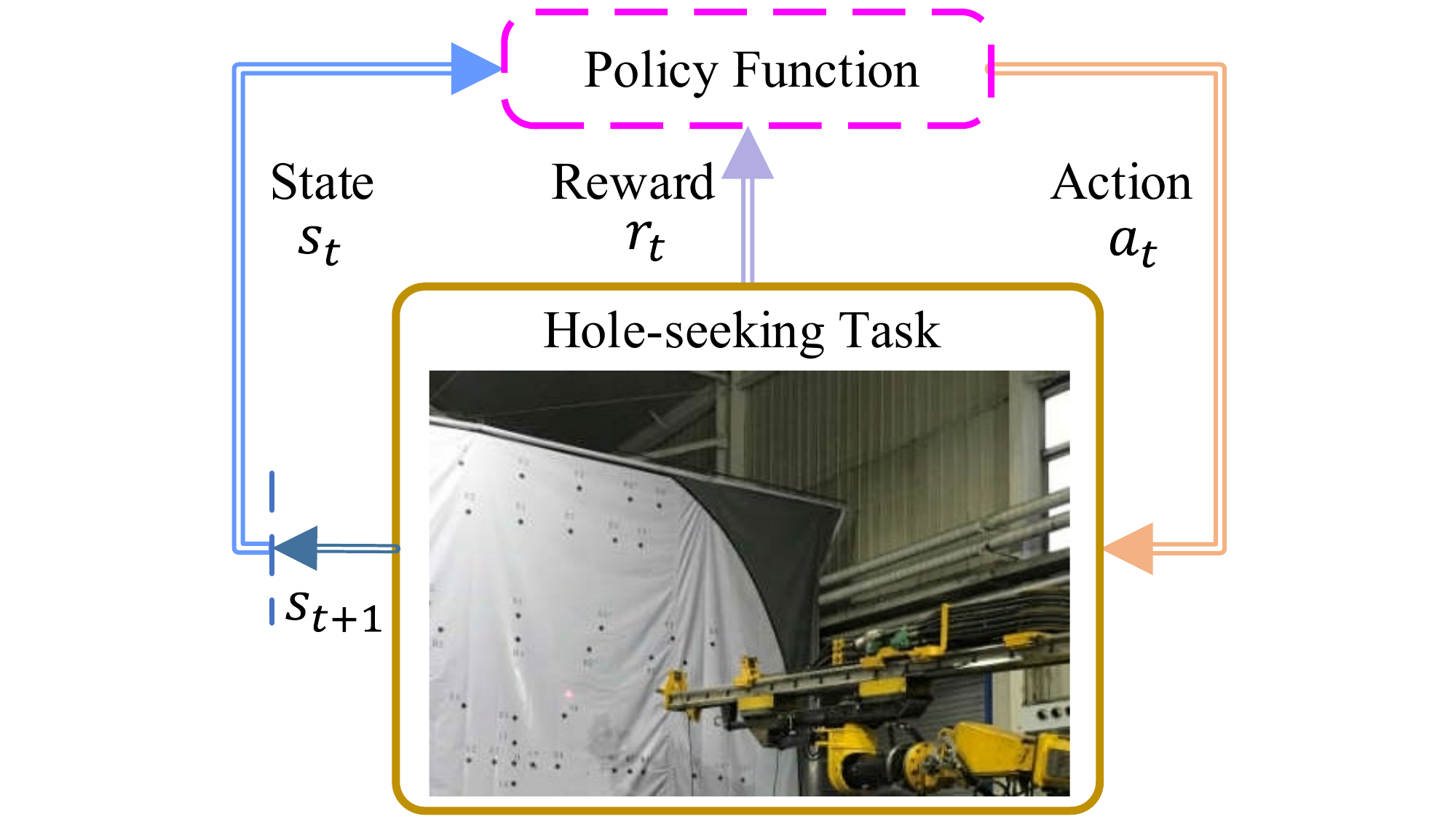}
\caption{RL-based policy training pipeline for drill boom hole-seeking.}  
\label{fig.RL_framework}
\end{figure}

Fig.~\ref{fig.RL_framework} presents a typical training pipeline for developing a drill boom control policy using RL. The policy function determines the action $a_{t}$ to be taken, based on the current state $s_{t}$, and receives the reward $r_{t}$ alongside the subsequent state $s_{t+1}$. In this section, we delve into the design of the state representation, action space, and reward function for this process.

\subsection{State representation design}
\label{subsec.state}
When considering the hole-seeking task, the state representation should consist of two kinds of information: drill boom system information and hole-seeking task information, symbolized as $s=[s_{\rm drill},s_{\rm task}]$. The drill boom system information provides a depiction of the spatial joint posture of the drill boom, which is independent of the task itself. On the other hand, the task information is task-specific and describes the relative positional relationship between the target hole and the actual drill end. 

\subsubsection{Drill boom system representation}
Existing research generally employs the Cartesian coordinates of each joint to represent joint posture\cite{li2021robot}. Each joint is characterized by a 3-dimensional coordinate vector, denoted by $(x,y,z)$. To describe the whole drill boom, 24-dimensional information is necessary given a total of eight joints, i.e., 
\begin{equation}
\label{eq.Cartesian}
s_{\rm drill}^{\rm Cartesian}=[x_1,y_1,z_1,x_2,y_2,z_2, \cdots ,x_8,y_8,z_8].
\end{equation}

However, obtaining the Cartesian coordinates for all joints is not straightforward. The coordinates of each joint are typically derived by solving forward kinematics using the DH method, relying on the joint angle and link length information. As the 3-dimensional Cartesian coordinates of all joints can be exclusively determined by the joint information in the DH coordinate, which depicts the positional relationship between adjacent joints, one natural idea is to directly represent the joint posture using DH parameters. By utilizing DH parameters to describe the drill boom system, it becomes easier to establish the mapping relationship between the joint motion and the resulting system state variation.

Since some DH parameters are constant depending on the structure of the drill boom, we only need to consider the variable parameters. For the boom displayed in Fig. \ref{fig.diagram}, each joint is either a prismatic pair or a revolute pair. For the prismatic pairs, i.e., joints 3 and 8, their angles are fixed, so only the length information is required, represented by $d_3$ and $d_8$. Similarly, for the other six revolute pairs, only the joint angle information is necessary, denoted as $\theta_{1,2,4,5,6,7}$. Therefore, the drill boom system can also be described as 
\begin{equation}
\label{eq.DH}
s_{\rm drill}^{\rm DH}=[\theta_1, \theta_2, d_3,\theta_4,\theta_5,\theta_6,\theta_7, d_8]. 
\end{equation}

Compared to the Cartesian representation, the DH representation possesses fewer dimensions. This often eases the learning challenge and improves the ultimate control performance. Given these advantages, we utilize $s_{\rm drill}^{\rm DH}$ as our preferred scheme for joint posture representation.

\subsubsection{Task information representation}
The role of the task information is to direct the drill end towards the expected position and direction. One simple approach to conveying this information is through the non-preview representation technique. This technique involves two components: the positional deviation and the angular deviation between the drill end and the start point of the target hole.

\begin{figure}[htbp] 
\centering 
\subfigure[Non-preview representation.]{
\includegraphics[width=0.40\textwidth]{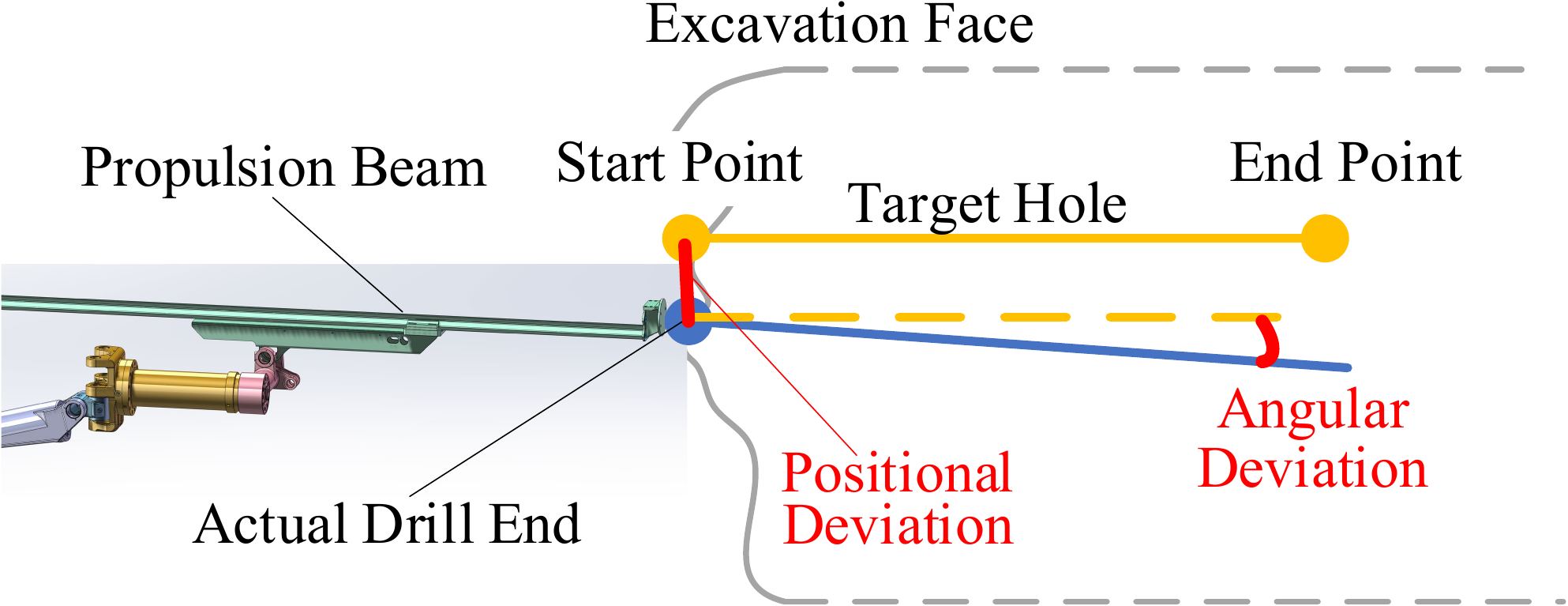}
\label{fig.Non-preview state}}
\quad
\subfigure[Preview representation.]{
\includegraphics[width=0.40\textwidth]{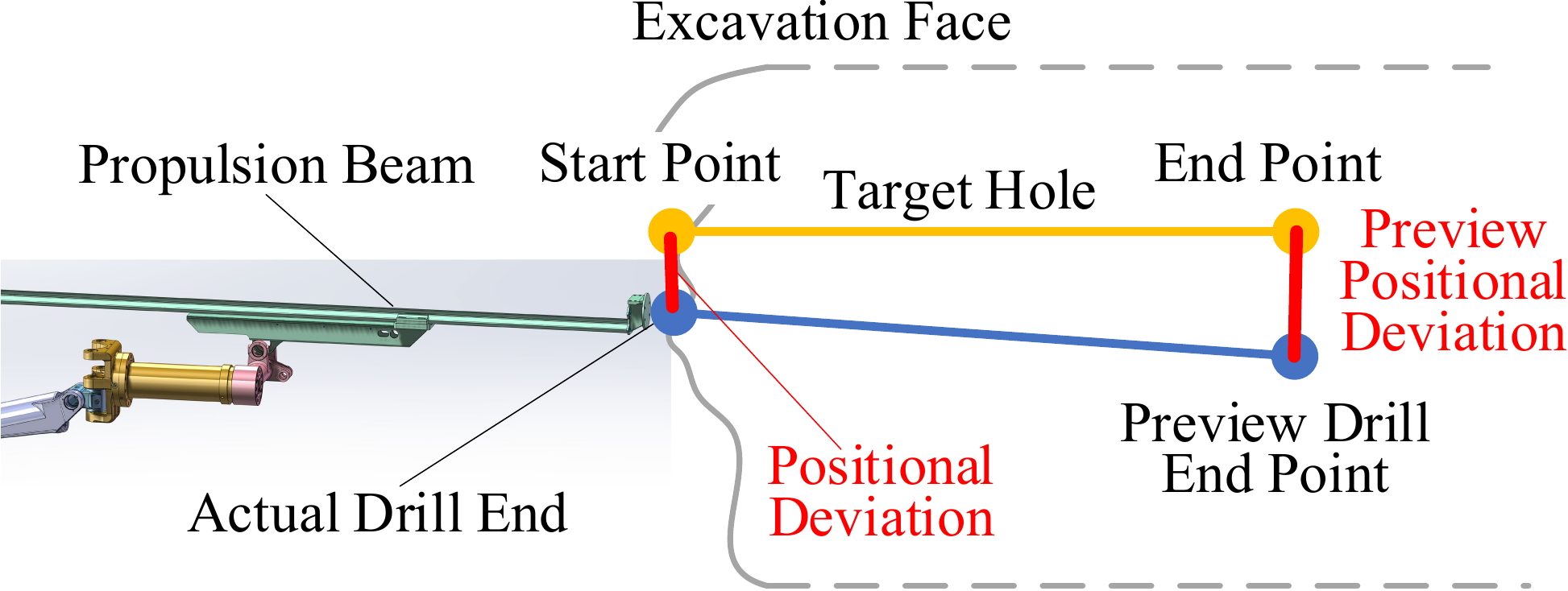}
\label{fig.Preview state}}
\caption{Non-preview and preview hole-seeking representation.}
\end{figure}

The positional deviation can be calculated by determining the disparity in Cartesian coordinates between the current drill end and the start point of the target hole:
\begin{equation}
\label{eq.start_gap}
\delta_{\rm current} = [x_{\rm drill}-\bar{x}_{\rm start}, y_{\rm drill}-\bar{y}_{\rm start}, z_{\rm drill}-\bar{z}_{\rm start}],
\end{equation}
where $(\bar{x}_{\rm start},\bar{x}_{\rm start},\bar{x}_{\rm start})$ represents the coordinate of the start point of the target hole.
The angular deviation can be measured by the difference between the direction vectors of the drill end and the target hole, represented as $\delta_{\rm angle}\in \mathbb{R}^3$. By combining the positional and angular deviation information in Fig.~\ref{fig.Non-preview state}, we can derive the non-preview representation of the task information:
\begin{equation}
\label{eq.NP-representation}
s^{\rm NP}_{\rm task}=[\delta_{\rm current}, \delta_{\rm angle}].
\end{equation}

In theory, if the learned policy could ensure that both $s_{\rm start}$ and $s_{\rm angle}$ are equal to zero, achieving high drilling accuracy throughout the entire process would not be difficult. However, slight positional and angular discrepancies are inevitable. In practical applications, even if $\|s_{\rm start}\|$ is relatively small, cumulative errors are likely to arise due to the large drilling depth (approximately 3m between the start and end points of the target hole). This generally results in a substantial gap between the preview drill end and the end point of the target hole. From the perspective of the entire drilling process, this approach can lead to subpar drilling performance. The fundamental reason behind this problem is that this characterization only considers the discrepancy between the drill end and the start point of the target hole, overlooking potential variations throughout the entire drilling process.

To address this issue, we propose a preview representation method to describe the information needed for hole-seeking, as depicted in Fig.~\ref{fig.Preview state}. Specifically, we replace the angular deviation $\delta_{\rm angle}$ in \eqref{eq.start_gap} with the preview positional deviation at the drill end point, denoted as $\delta_{\rm preview}$. The preview positional deviation is determined by the difference between the end point of the target hole and the preview drill end point along the extension line of the propulsion beam, which is expressed as
\begin{equation}
\label{eq.end_gap}
\delta_{\rm preview} = [x_{\rm preview}-\bar{x}_{\rm end}, y_{\rm preview}-\bar{y}_{\rm end}, z_{\rm preview}-\bar{z}_{\rm end}],
\end{equation}
where $(x_{\rm preview},y_{\rm preview},z_{\rm preview})$ are the coordinates of the preview point of the drill end and $(\bar{x}_{\rm end},\bar{y}_{\rm end},\bar{z}_{\rm end})$ represent the coordinates of the end point of the target hole.
Subsequently, we substitute $\delta_{\rm angle}$ in \eqref{eq.start_gap} with \eqref{eq.end_gap} to obtain the preview representation:
\begin{equation}
\label{eq.P-representation}
s^{\rm P}_{\rm task}=[\delta_{\rm current}, \delta_{\rm preview}].
\end{equation}
By incorporating the preview drilling discrepancy into the state representation, one can more effectively ensure the accuracy of the entire drilling process.

Based on the aforementioned analysis, this study combines the joint posture state characterized by DH coordinates and the task  representation based on preview drilling discrepancy to form the final state space for RL:
\begin{equation}
\label{eq.state-representation}
s=[s_{\rm drill}^{\rm DH},s^{\rm P}_{\rm task}].
\end{equation}

\subsection{Action selection}

In terms of motion execution, we choose to use an incremental update mechanism. In particular, we choose the change rate of joint posture, which corresponds to the change rate of DH parameters in \eqref{eq.DH},  as the action input at each time step, denoted as 
\begin{equation}
\label{eq.action}
a\doteq \Delta s_{\rm drill}^{\rm DH}.
\end{equation}
In this case, the relationship between joint postures at successive moments is
\begin{equation}
\label{eq.state_transition}
s_{{\rm drill},t+1}^{\rm DH}=s_{{\rm drill},t}^{\rm DH}+a_t/f,
\end{equation}
where $f$ represents the control frequency, set to $10$Hz in this study. Taking into account the inherent limitations of mechanical mechanisms, we assume that
\begin{equation}
\label{eq.action space}
a\in [a_{\rm min},a_{\rm max}],
\end{equation}
where $a_{\rm min}$ and $a_{\rm max}$  depend on the practical capability of the actuator. In this study, we set $a_{\rm max}=[0.08, 0.08, 20, 0.08, 0.08,0.12,0.08, 20]$ and $a_{\rm min}=-a_{\rm max}$. Please note that the action unit for the revolute pair is rad/s, while it is mm/s for the prismatic pair.

\subsection{Reward function design}
Based on the state and action mentioned above, the reward function is designed as
\begin{equation}
r_t = -\omega _1 \vert|\delta_{{\rm start},t}\vert| - \omega _2 \vert|\delta_{{\rm preview},t}\vert| - \omega_3 \vert|a_t\vert|,
\label{eq.reward}
\end{equation}
where $\omega _1$, $\omega_2$, and $\omega_3$ are weights used to balance the importance of different terms. The former two terms penalize the hole-seeking error, promoting alignment between the actual drilling trajectory and the target hole. The last term regularizes the action to ensure control smoothness. In this study, we set $\omega = [3, 3, 0.005]$.

\section{Experiments}
\label{sec.experiment}

This section conducts simulated experiments to validate the efficacy of the proposed RL-based integrated drill boom control method.

\subsection{Experimental details}

\begin{figure}[htbp] 
\centering 
\includegraphics[width=0.4\textwidth]{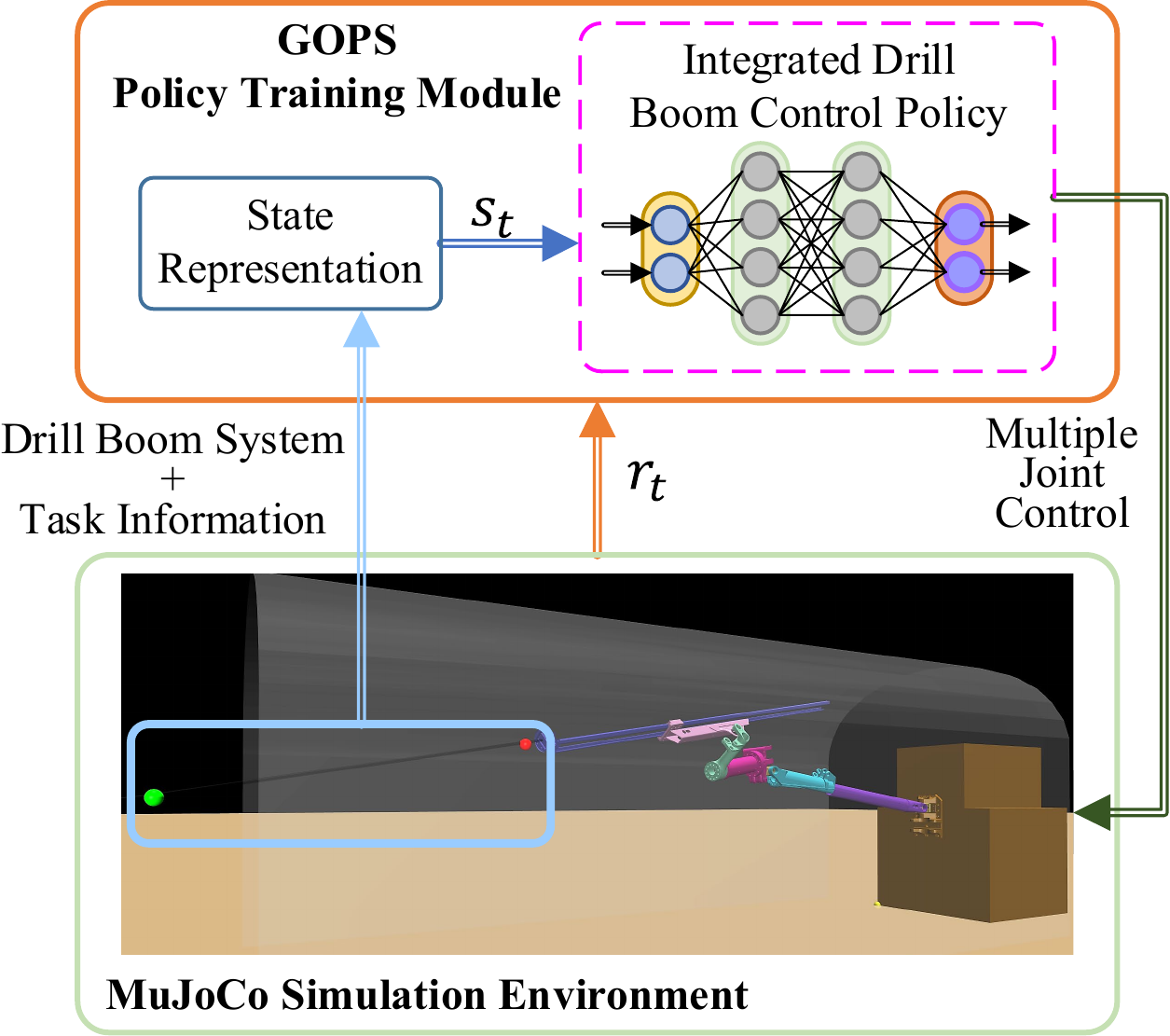}
\caption{\small Policy learning based on simulation environment.}
\label{fig.co-simulation}
\end{figure}

Utilizing  the MDP elements delineated in Section \ref{sec.MDP hole-seeking}, we can directly apply RL techniques to learn a parameterized integrated drill boom control policy. This study utilizes the General Optimal control Problem Solver (GOPS) \cite{wang2023gops}, which incorporates numerous mainstream RL algorithms, for policy learning. Specifically, four RL algorithms suited for continuous control settings are selected, including DSAC\cite{duan2021distributional, duan2023dsact}, SAC\cite{haarnoja2018soft}, TD3\cite{Fujimoto2018TD3}, and DDPG\cite{lillicrap2015DDPG}. To facilitate data interaction and visual verification during policy learning, we build a hole-seeking task simulation environment based on the MuJoCo platform\cite{todorov2012mujoco}, as depicted in Fig.~\ref{fig.co-simulation}. We employ multilayer perceptrons (MLPs) to represent both the value function and policy function. Each MLP comprises two hidden layers with 256 neurons per layer. The Adam optimizer, with a learning rate of $0.001$, is utilized to update both value and policy networks.

\subsection{Results}

\subsubsection{Overall Performance}
We run each RL algorithm five times with different random seeds. To assess the hole-seeking accuracy of the learned policy, we define two criteria: the current hole-seeking error $\epsilon_{\rm current}$  and the preview hole-seeking error $\epsilon_{\rm preview}$, defined as 
\begin{equation}
\epsilon_{\rm current}=\mathbb{E}(\|\delta_{\rm current}\|), \ \epsilon_{\rm preview}=\mathbb{E}(\|\delta_{\rm preview}\|).
\end{equation}
These two errors could provide a comprehensive reflection of the accuracy throughout the entire drilling process.

\begin{figure}[!htbp] 
\centering 
\includegraphics[width=0.45\textwidth]{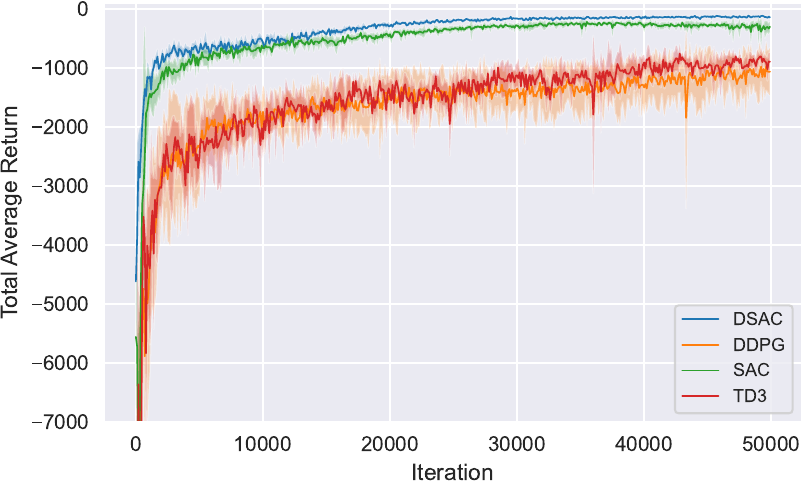}
\caption{Learning curves. The solid line shows the mean value over five runs, and the shaded area represents the 95$\%$ confidence interval. }
\label{fig.algorithm}
\end{figure}

The learning curves and final performance results are exhibited in Fig.~\ref{fig.algorithm} and Table \ref{tab.final performance}, respectively. Results demonstrate that all four algorithms can converge to a relatively stable performance within $5\times 10^4$ iterations. Relatively, DSAC significantly outperforms the other three algorithms in terms of both final return and hole-seeking accuracy. Specifically, the hole-seeking accuracy achieved by DSAC fully complies  with practical demands.  This provides numerical evidence  supporting the effectiveness of the proposed RL-based integrated drill boom control method. In other words, based on the proposed integrated drill boom control framework and MDP formulation, we can select an appropriate RL algorithm to discover a competent hole-seeking control policy. 

\begin{table}[!htp]
\renewcommand{\arraystretch}{1.5}
\setlength{\tabcolsep}{5pt}
\centering 
\caption{Final performance comparison.}
\begin{tabular}{lccl}
\toprule
Algorithm  & Final Average Return      & $\epsilon_{\rm current}$ [mm] & $\epsilon_{\rm preview}$ [mm] \\ 
\midrule
DSAC       & \textbf{-130.237$\pm$22.083}   & \textbf{4.291$\pm$1.882}       & \textbf{6.263$\pm$2.588}               \\
SAC        & -328.303$\pm$88.394           & 31.530$\pm$14.999       & 33.948$\pm$12.892               \\
TD3        & -885.038$\pm$149.365           & 33.774$\pm$17.127       & 40.557$\pm$12.399              \\
DDPG       & -1057.722$\pm$372.203          & 35.937$\pm$18.997       & 54.441$\pm$18.330               \\
\bottomrule
\end{tabular}
\label{tab.final performance}
\end{table}

\begin{figure}[htbp] 
\centering 
\makeatletter
\renewcommand{\@thesubfigure}{\hskip\subfiglabelskip}
\makeatother
\vspace{-0.1in}
\subfigure[]{
\includegraphics[width=0.22\textwidth]{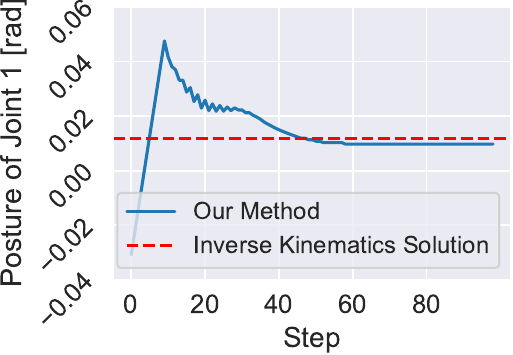}}
\vspace{-0.28in}
\subfigure[]{
\includegraphics[width=0.22\textwidth]{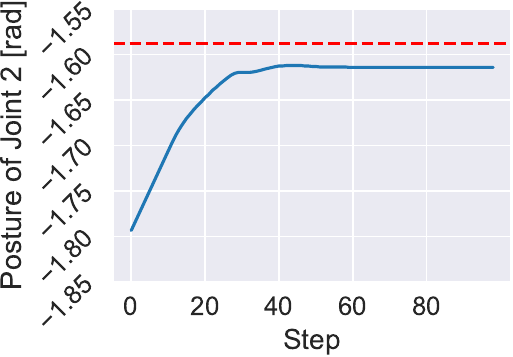}}
\vspace{-0.1in}
\subfigure[]{
\includegraphics[width=0.22\textwidth]{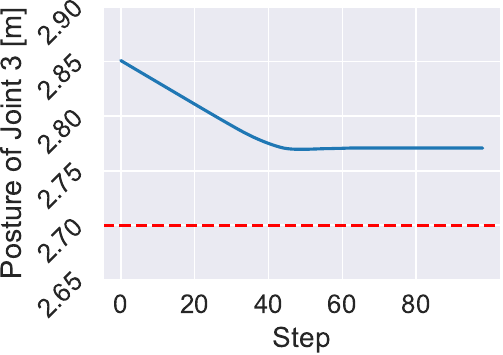}}
\vspace{-0.2in}
\subfigure[]{
\includegraphics[width=0.22\textwidth]{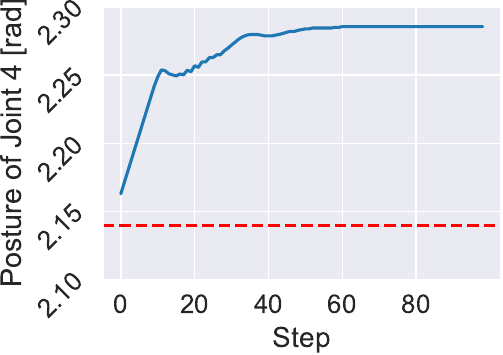}}
\vspace{-0.15in}
\subfigure[]{
\includegraphics[width=0.22\textwidth]{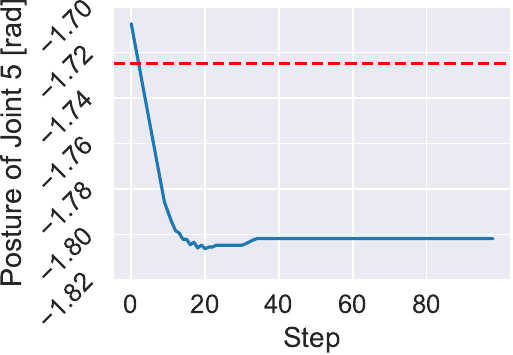}}
\vspace{-0.1in}
\subfigure[]{
\includegraphics[width=0.22\textwidth]{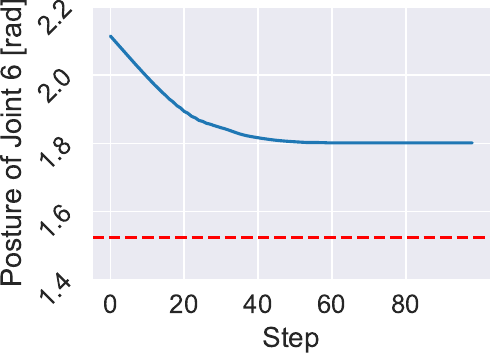}}
\subfigure[]{
\includegraphics[width=0.22\textwidth]{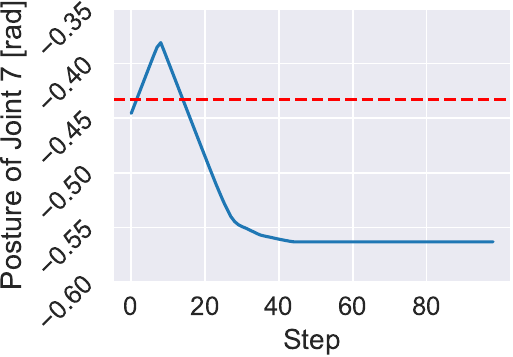}}
\subfigure[]{
\includegraphics[width=0.22\textwidth]{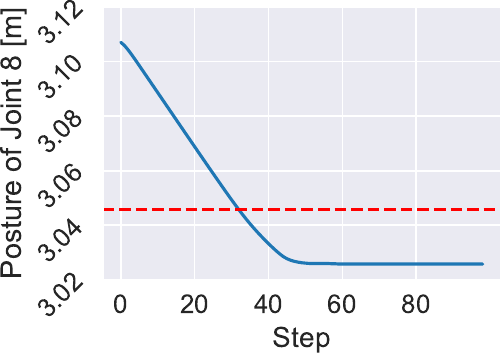}}
\caption{The control curve employing our method is depicted by the blue lines, and the red dotted lines portray the inverse kinematics solution corresponding to the same target hole. Owing to the non-uniqueness of inverse kinematics solutions, the control curves converge towards a substantially distinct value compared to the red lines.}
\label{fig.joint_posture}
\end{figure}

\begin{figure}
    \centering\includegraphics[width=0.4\textwidth]{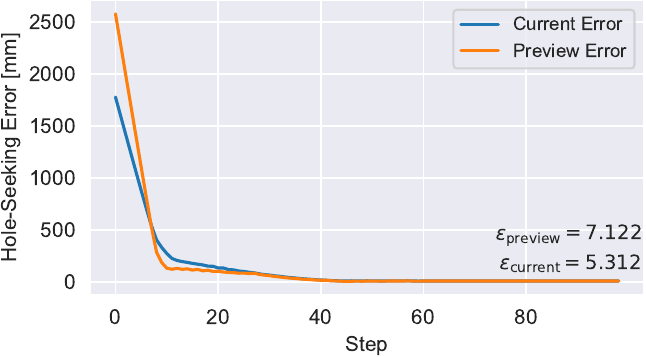}
    \caption{Current and preview errors curves in a hole-seeking process.}
    \label{fig.simu_error}
\end{figure}

The blue curves in Fig. \ref{fig.joint_posture} intuitively display the control process of joint posture in a hole-seeking task, with the corresponding hole-seeking error curves illustrated in Fig. \ref{fig.simu_error}.
It is apparent that the integrated control can quickly guide the end of the drill boom to converge to the drilling target with a significantly low error.

\subsubsection{Time efficiency} The proposed integrated drill boom control method differs from the traditional hierarchical drill boom method in both the decision-making and control processes. For decision-making, our method substitutes the inverse kinematics solution with multi-step decision-making based on the learned policy. Regarding the control process, our method simultaneously controls multiple joints as opposed to in a sequential manner. Next, we employ the policy derived from DSAC to assess the time efficiency of our method, specifically in relation to the decision-making and control processes.

To analyze the influence of different control mechanisms on hole-seeking efficiency, we conducted 100 simulations and counted the actual effective steps taken by each joint to reach the final posture during the integrated hole-seeking process. The maximum number of moving steps across eight joints corresponds to the number of control steps for the integrated drill boom control. The aggregate of the motion steps for all eight joints can be approximately deemed as the control steps needed for the hierarchical control framework. Table \ref{tab.execution steps} enumerates the control steps obtained through this analysis. Given the same initial and final joint posture, the control steps required by the hierarchical method are about 5.7 times that of the integrated method. This indicates that the integrated control method can greatly enhance control efficiency by coordinating multiple joints simultaneously. 
\begin{table}[!htp]
\centering 
\renewcommand{\arraystretch}{1.5}
\setlength{\tabcolsep}{5pt}
\caption{Execution steps for hole-seeking task.}
\begin{tabular}{llllll}
\toprule
             & Integrated Control & Hierarchical Control     \\
\midrule
Step number & \textbf{299.631$\pm$7.440}         &  1708.157$\pm$37.007     \\ 
\bottomrule
\end{tabular}
\label{tab.execution steps}
\end{table}

According to the simulation results, the policy network takes approximately 0.318ms to generate single-step control inputs for all joints. On the basis of the average control steps needed for the hole-seeking task as outlined in Table \ref{tab.execution steps}, the total time consumption for the decision-making process under the integrated method is roughly 97.02ms. This is considerably less compared to existing numerical inverse kinematic solvers, which generally take between 0.4s to 0.6s\cite{sugihara2011solvability,husty2007new}. In addition, since the integrated method optimizes the control policy by maximizing the accumulated reward, the learned policy typically identifies the optimal final joint posture requiring the least number of control steps. However, as inverse kinematics do not  yield a unique solution, it is challenging for the hierarchical method to find the expected joint posture with the minimum control steps. 
As evidenced by the example illustrated in Fig. \ref{fig.joint_posture}, our method comprehensively considers the impact of joint actions on execution efficiency, and adopts coordinated motions that are completely different from the inverse kinematics solution, despite corresponding to the same target hole.
Therefore, in practical operation, the control efficiency of the integrated method generally surpasses that of hierarchical control by more than 5.7 times, given that the integrated method can use a shorter distance to reach the target point.

\subsubsection{Ablation studies}
\begin{figure}[htbp] 
\centering 
\subfigure[Current hole-seeking error.]{
\includegraphics[width=0.45\textwidth]{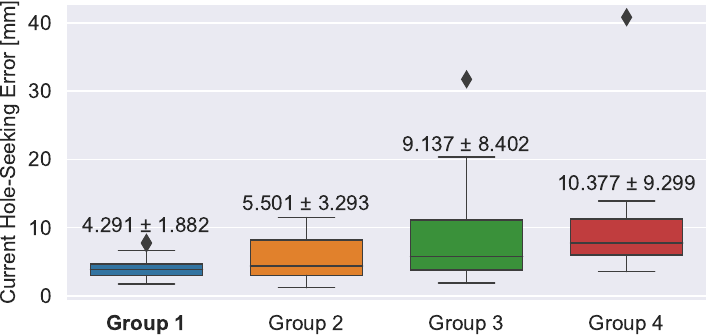}
\label{fig.Current hole-seeking error}}
\quad
\subfigure[Preview hole-seeking error.]{
\includegraphics[width=0.45\textwidth]{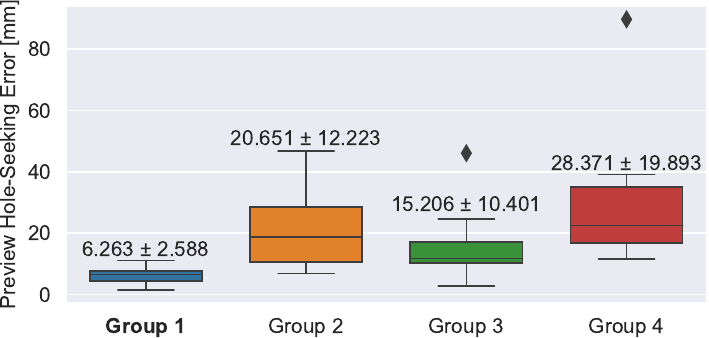}
\label{fig.Preview hole-seeking error}}
\caption{Performance of different state representations.}
\label{fig.error}
\end{figure}

In this section, we carry out ablation studies to validate the effectiveness of the state representation described in \eqref{eq.state-representation}. Four different state representations are considered, as detailed in Table \ref{tab.representation combinations}. We run DSAC based on each representation five times, and compare the average hole-seeking error over 100 random simulations, as illustrated in Fig. \ref{fig.error}. The results indicate that our state representation, employing the DH joint representation and preview hole-seeking information, yields the highest accuracy throughout the entire drilling process.

\begin{table}[!htp]
\renewcommand{\arraystretch}{1.5}
\centering 
\caption{State representation baselines.}
\begin{tabular}{cc}
\toprule
Group & State representation \\
\midrule
1 (our method)     & $s_{\rm drill}^{\rm DH}$ in \eqref{eq.DH} $+$$ s^{\rm P}_{\rm task}$ in \eqref{eq.P-representation}            \\
2     & $s_{\rm drill}^{\rm DH}$ in \eqref{eq.DH} $+$ $ s^{\rm NP}_{\rm task}$ in \eqref{eq.NP-representation}         \\
3     & $s_{\rm drill}^{\rm Cartesian}$ in \eqref{eq.Cartesian} $+$ $ s^{\rm P}_{\rm task}$ in \eqref{eq.P-representation}          \\ 
4     & $s_{\rm drill}^{\rm Cartesian}$ in \eqref{eq.Cartesian} $+$ $ s^{\rm NP}_{\rm task}$ in \eqref{eq.NP-representation}        \\
\bottomrule
\end{tabular}
\label{tab.representation combinations}
\end{table}

On one hand, the preview mechanism, which mitigates error accumulation in the drilling process by simultaneously constraining the current and preview disparities between the drill and target hole, improves both the current and preview hole-seeking accuracy for Group 1 (and Group 3) as compared to Group 2 (and Group 4). To be specific, judging from the performance of Groups 1 and 2, the preview representation reduces the current and preview errors by 22.0\% and 69.7\%, respectively, in contrast to the non-preview representation.

On the other hand, Group 1 (and Group 2) exhibits superior seeking accuracy for both current and preview points in comparison to Group 3 (and Group 4). This suggests that the DH representation of joint posture can enhance the overall control performance, owing to its lower dimensionality. Specifically, when considering the performance of Groups 1 and 3, the DH joint representation reduces the current and preview errors by 53.1\% and 58.8\%, respectively, compared to the Cartesian joint representation.

In summary, the proposed integrated drill boom control method delivers superior performance in both hole-seeking accuracy and time efficiency.  This suggests that our method has significant potential for real-world application, as it can enhance drilling efficiency while simultaneously mitigating common issues such as overbreak and underbreak caused by low hole-seeking accuracy.

\section{Conclusion}
\label{sec.conclusion}
This study presented an RL-based integrated control method for drill boom hole-seeking control. We formulated the hole-seeking task as an MDP and employed RL to learn an offline policy network, which directly outputs multi-joint control signals based on the state representation. The state includes the joint posture in the DH coordinate system and the discrepancy between the drill end and target hole in both current and preview drill points. This greatly enhances the hole-seeking accuracy compared to representations based on Cartesian coordinates and non-preview hole-seeking information. By utilizing the DSAC algorithm, the learned policy has achieved extremely high hole-seeking accuracy, keeping both the current and preview hole-seeking errors within 1cm. Notably, our method shows significant improvements (about 5.7 times) in terms of execution efficiency when compared to the hierarchical method. Please note that this study mainly focuses on hole-seeking accuracy and time efficiency. In future work, we will further consider state constraints to ensure the safety of the hole-seeking process. 

\section*{Acknowledgment}
\label{sec.acknowledgement}
We extend our profound acknowledgment to Zhangjiakou Xuanhua Huatai Mining and Metallurgical Machinery Co., Ltd. for their funding and invaluable suggestions for this study.



\end{document}